\documentclass[runningheads]{llncs}
\usepackage{graphicx,color,xcolor} 
\usepackage[]{algorithm2e}
\usepackage{amsmath}
 \usepackage{multirow}
\graphicspath{{Figures/}}

\DeclareMathOperator{\argmin}{arg\,min}

\SetKwInput{KwData}{Input}

\begin{document}
\title{REBEC: Robust Evolutionary-based Calibration Approach for the Numerical Wind Wave Model}
\titlerunning{REBEC: Calibration Approach for the Numerical Wind Wave Model}
%
\author{Pavel Vychuzhanin\inst{1} \and
Nikolay O. Nikitin\inst{1} \and
Anna V. Kalyuzhnaya\inst{1}}
\authorrunning{Pavel Vychuzhanin et al.}
%

\institute{ITMO University, 49 Kronverksky Pr. St. Petersburg, 197101, Russian Federation
\email{\{pavel.vychuzhanin, nikolay.o.nikitin\}@gmail.com}
}

%
\maketitle              
\begin{abstract}
The adaptation of numerical wind wave models to the local time-spatial conditions is a problem that can be solved by using various calibration techniques. However, the obtained sets of physical parameters become over-tuned to specific events if there is a lack of observations. In this paper, we propose a robust evolutionary calibration approach that allows to build the stochastic ensemble of perturbed models and use it to achieve the trade-off between quality and robustness of the target model. The implemented robust ensemble-based evolutionary calibration (REBEC) approach was compared to the baseline SPEA2 algorithm in a set of experiments with the SWAN wind wave model configuration for the Kara Sea domain. Provided metrics for the set of scenarios confirm the effectiveness of the REBEC approach for the majority of calibration scenarios.

\end{abstract}

\section{Introduction}

The various tasks of offshore development and coastal shipping make it necessary to use the regional configurations of the numerical wind wave models to reproduce historical extreme events and predict potential hazards. To obtain the forecasts and hindcasts of desired quality, the suitable physical parameters of models should be identified for the specific simulation conditions.

The numerical model calibration of ocean wind wave model involves the fitting of simulation results with the in-situ and satellite wave measurements. The purpose of calibration is the identification of the physical parameters set that allows minimising the discrepancy between the model and observations. 

However, it is a sophisticated task to calibrate the model manually even with the metocean experts' involvement. The modern wind wave models are computationally intensive, and each simulation run can take hours to compute. Also, a dramatically low time-spatial coverage of the available historical wave measurements and low quality of atmospheric reanalyses in some regions (like the Arctic seas, in particular the Kara Sea region described in the paper) makes it hard to validate the parameter set reliably. The obtained parameters with minimal discrepancy can be very specialised in case of over-fitting to the low number of observed data points and can actually decrease the quality of long-term simulation results in non-observed locations or time ranges \cite{brynjarsdottir2014learning}.

There are many well-known optimisation approaches that can be applied to automate the parameters' tuning for environmental models as well as \cite{hourdin2017art}. Despite this, in the paper a task-specific robust evolutionary algorithm is proposed. It allows to make reliable calibration decisions in situations with high environmental uncertainty and tries to ensure a tolerable solution identification.

At the moment, the modern atmospheric reanalysis still has quality issues in the Arctic region \cite{fredriksen2018evaluation}. We proposed an algorithm that establishes artificial diversity for wind velocity fields. It was used to generate the probabilistic ensemble of input wind fields to take the impact of the surface forcing uncertainty into account. Then, the multi-objective fitness function was used to achieve the trade-off between robustness and performance of the optimised model. 

We conducted a set of experiments to verify the effectiveness of the proposed approach against the baseline SPEA2 algorithm using the Kara Sea domain and the SWAN (Simulating WAves Nearshore)\cite{booij1999third} model as the case study. The nine spatially distributed points were chosen to analyse the performance and robustness of the model's configurations obtained after calibration in one-month training runs of the model. The several configurations with different subsets of calibration and validation points were compared to estimate the statistical metrics of optimisation effectiveness for both algorithms. 

This paper is structured as follows. Sec.~\ref{sec_problem} describes the problem statement and mathematical formalisation of the robust optimisation task. Sec.~\ref{sec_related} provides an overview of various calibration approaches and their applicability for the problem. Sec.~\ref{sec_evo_alg} contains a detailed description of the baseline SPEA2 algorithm and the proposed robust algorithm. Sec.~\ref{sec_experiments} is dedicated to the experimental studies (model configuration, datasets, results and metrics). Sec.~\ref{sec_concl} summarises the obtained results and highlights of the key findings.

\section{Problem statement}
\label{sec_problem}

As it was noted in introduction, coverage of observed met-ocean data (especially oceanic observations) is extremely sparse. Although, reliable information about met-ocean characteristics is needed in many regions (e.g. Arctic seas). That's why during last decades it became a common practice to obtain the information about met-ocean events and processes from forecasting or hindcasting (retrospective) simulation results from numerical hydrodynamic models. Nevertheless, for solving such task the numerical models should be fitted (through model parameters) to the certain water area. Taking into account few spatial points and small sizes of datasets with observations, there is a serious risk of model overfitting when model fits to specific features of observed data instead of fitting to common features of the target region. Description of the solution to this problem is the main goal of this article. 

Hydrodynamic model fitting through the tuning of model parameters (or model calibration) can be formulated as an optimisation task. For this purpose, it is reasonable to present the simulation process in a general mathematical notation (\ref{eq_sim_proc}).

\begin{equation}
\label{eq_sim_proc}
Y = \{Y_1, Y_2, ..., Y_k\} = M(\xi\mid\theta), 
\end{equation}
where \( Y = \{Y_1, Y_2, ..., Y_k\} \) denotes multivariate output data (simulated fields, e.g. wave heights), \( M(\bullet) \) is the model operator, \( \xi \) is the input data (boundary and initial conditions), \( \theta \) is the set of model parameters. 

With that, the tuning of model parameters (or model calibration) can be formalized in terms of multi-objective optimisation  in the model parameter space and written as:

\begin{equation} \begin{split}
\label{eq_opt}
& \theta_{opt} = \argmin_{\theta}{F(\theta)}, \\
& F(\theta) = \mathcal{G}(f_i(\theta, Y, \{x,y\})), \\
\end{split} \end{equation}
where \( \mathcal{G}(\bullet) \) is an operator for multiobjective transformation to \( F\) , \( f_i\) is the objective function, i = 1...n , \(\{x,y\}\) are spatial coordinates of a point-of-interest.

In a case of wind waves hindcasting, the poor time and spatial coverage of observations make the model optimisation  much harder. The over-fitting of the solution to the specific events represented in small data samples can cause a non-optimal model configuration with lower quality under different external conditions. One of the ways to improve the robustness of optimisation  results is to enlarge training dataset with new instances with relatively small artificial disturbances. This issue makes it necessary to take the simulation uncertainty factors into account. 

The uncertainty in the wind wave model can be represented not only by disturbances in design variables \cite{paenke2006efficient}. There are deviations in the environment variables that can be represented through input data sets diversity (for the SWAN model the wind forcing obtained from atmospheric reanalysis is most important). In this case input data \( \xi \) should be transformed to ensemble realisation  \( \{\xi\}_n  = \{\xi_1, ..., \xi_n\} \) by addition of artificial disturbance (or noise) and equation (\ref{eq_opt}) transforms into equation (\ref{eq_opt_ens}). A detailed description of the ensemble procedure is given in Sec. 4.3. 

\begin{equation}
\label{eq_opt_ens}
\begin{array}{cc}
 \theta_{rob} = \argmin_{\theta}{\tilde{F}(\theta\mid\{\xi\}_n)}, \\
 \tilde{F}(\theta\mid\{\xi\}_n)  = \mathcal{G}(\tilde{f_i}(\theta\mid\{\xi\}_n,Y, \{x,y\})). \\
\end{array} 
\end{equation}

An ensemble objective function \(\tilde{f_i}\) defines landscape of objective function over the space of parameters considering ensemble of input states \( \{\xi\}_n \). As an example, ensemble fitness function can be represented by the expected function for the ensemble of runs with small disturbances in input data (shown in equation (\ref{ens_fitness_exp})).  This approach can be used to produce better solutions for the set of diverse environmental scenarios and increase the expected performance.
\begin{equation}
\label{ens_fitness_exp}
\tilde{f}(\theta\mid\{\xi\}_n) = \int _ { - \infty } ^ { \infty } f ( \mathbf { x },\mathbf { \xi } + \boldsymbol { \delta } ) \cdot p ( \boldsymbol { \delta } ) d \boldsymbol { \delta }
\end{equation}

As an example of the hydrodynamic model for experimental studies, third-generation wind wave model SWAN \cite{booij1999third} was chosen. 
The wind waves are surface waves in the oceans and seas that caused by the interaction between water masses and sea-level wind. Wind waves models of third-generation (e.g. SWAN) allow to simulate the wave spectra and to reconstruct characteristics of waves (e.g. heights, periods, directions). The SWAN model can be described with the action balance equation (\ref{eq_act_bal}).

\begin{equation}
\label{eq_act_bal}
\frac {\partial} {\partial t} N + \frac {\partial} {\partial x} c_x N + \frac {\partial} {\partial y} c_y N + \frac {\partial} {\partial \sigma} c_\sigma N + \frac {\partial} {\partial \theta} c_\theta N = \frac{S} {\sigma},
\end{equation}
where on the left-hand side \(N = \frac{E} {\sigma}\) denotes the wave action density and \(E\) is an energy of wave spectrum, \(\sigma\) is the relative frequency, \(\theta\) is the group wave direction, \(c\) is the group velocity in corresponding space. The right-hand side represents the source and sink term in a form equation (\ref{eq_form}).

\begin{equation}
\label{eq_form}
S = S_{in} + S_{ds} + S_{nl}, 
\end{equation}
where \(S_{in}\) is the input energy obtained by wind, \(S_{ds}\) is the energy of dissipation and \(S_{nl}\) denotes the energy of wave-wave nonlinear interaction.

These three terms represent the genesis of wave energy sources/sinks and are a powerful handle for wave model fitting. From this point of view, it is convenient to express energy sources through model parameters. Wind energy is characterised by the drag function (DRG), wave dissipation --- by the wave breaking (STMP) and bottom friction (CFW) functions. Energy flow from nonlinear interactions is relatively small and wasn't taken into account in the current paper.

In the frame of this article, the experimental study (Sec.~\ref{sec_experiments}) was provided to assess the practical effectiveness of the proposed robust calibration method in comparison with the general-purpose calibration algorithms. The SWAN model configuration for the Kara Sea was chosen as a case study because of value of this region for offshore industrial development and extremely low density of sensors in areas of interest.

\section{Related work}
\label{sec_related}

Model calibration or tuning is a subject with extensive literature \cite{williams2017guidance,hourdin2017art}. The conservative approach is to estimate the parameters in an expert way \cite{jin2001calibration,mortlockcalibration}. It includes the development of several candidate sets of parameters based on previous simulation experience and manual individual adjustment of each parameter. The quality metrics for the model quality assessment are calculated with the comparison of model time series and historical values obtained from the reanalyses and observations.

Since the manual "trials-and-errors" method is time-consuming and gives solution only for particular model setup, the automatic calibration of models is widely used for different aspects of environmental simulations like atmospheric \cite{duan2017automatic} and ocean \cite{williamson2017tuning} forecasting tasks. As a basic approach, the space-filling design for the parameter space can be used \cite{wainwright2005environmental} for model calibration. However, the high-resolution configurations of wind wave models of 3rd generation are computationally-intensive and require a lot of time to process the appropriate date range and spatial domain. This problem makes it necessary to reduce the number of runs required for calibration.

There are many well-known optimisation methods applied to environmental models like derivative-free optimisation \cite{van2003calibration}, various Bayesian optimisation methods \cite{cornejo2018bayesian} and surrogate-assisted methods \cite{james2018machine}.

However, the evolutionary (genetic) algorithms are efficient enough to perform a robust solution search \cite{liu2007automatic} in a complex parameter space with a lack of historical data for quality assessment \cite{schmitt2015half}. The applicability of evolutionary algorithms for SWAN wave model calibration is demonstrated in \cite{kovalchuk2018conceptual}.

The robust optimal design approaches have a lot of applications in many fields \cite{zang2005review}. They are often based on Monte Carlo methods that allow representing the uncertainty from different sources \cite{che2014monte}. The perturbation-based ensemble allows sampling the modelling uncertainty in a more systematic way \cite{rougier2009analyzing}. A set of simulation with small differences induced by stochastic modifications allows to increase the variability of the calibration dataset and improve the quality of models \cite{o2018ensemble}.

Nevertheless, the discrepancy usually simulated as additional noise in model output and observations\cite{bhat2010computer} without taking the actual sources of external uncertainty (e.g. wind forcing for reanalysis) into account. The task of a reliable calibration of a wave model for a specific domain with poor observational coverage makes it necessary to implement the approach that combines the ensemble-based diversity of environmental variables with multi-objective evolutionary optimisation.  

\section{Evolutionary algorithms for models fitting}
\label{sec_evo_alg}

We compared the robust wave model calibration with a baseline solution --- the multi-objective evolutionary algorithm that estimates the most suitable solution without taking uncertainty into account. The other approach is based on the same algorithm with modified fitness functions --- it estimates the performance and robustness of the solution with the ensemble of forecasts obtained from several model runs with noised inputs. The source code of both algorithms was implemented in Python and available in \cite{github-sources}.

\subsection{Baseline approach}

The commonly used SPEA2 multi-objective optimisation algorithm \cite{zitzler2001spea2} was chosen as a baseline solution for the calibration task. In terms of evolutionary algorithms, in our case, each individual corresponds to a genotype represented by a certain set of model parameter and the phenotype (values of the objective function) are the errors of the model predictions, corresponding to these parameters. At each iteration of evolution, the Pareto-optimal set of individuals is selected according to the values of the fitness function, and all non-dominated solutions are saved in the archive. Then the mating pool is filled with a binary tournament selection and recombination and a mutation operator are applying for each individual. The resulting mating pool becomes a new population at the next iteration of the algorithm.

Despite the fact that some modern evolutionary algorithms outperform SPEA2 in some synthetic tasks \cite{li2019empirical}, we decided to base the experiments on a well-studied \cite{kovalchuk2018conceptual} algorithm to separate the impact from the proposed ensemble-based modifications from other features' influence. 

\subsection{Robust ensemble-based evolutionary calibration (REBEC) approach}

The main disadvantage of the baseline algorithm is that the model variables optimise exactly for the specific conditions that were used for the fitness function evaluation. It allows to maximise the performance for the observed case, but the solution found can be unstable even after small changes in external conditions. The lack of the time-spatial coverage of observational data for wave parameters in target regions makes it complicated to take the different external uncertainties (e.g. forcing-induced, resolution-induced, etc) into account.

The more robust approach to model parameters optimisation can be implemented using the ensemble of wave models configured using different input data sets. We can form the stochastic ensemble of wave models with the perturbed wind forcings and search for more robust model parameters using this ensemble instead of a single model with certain forcing.

\begin{figure}
\centering
\includegraphics[scale=2.0]{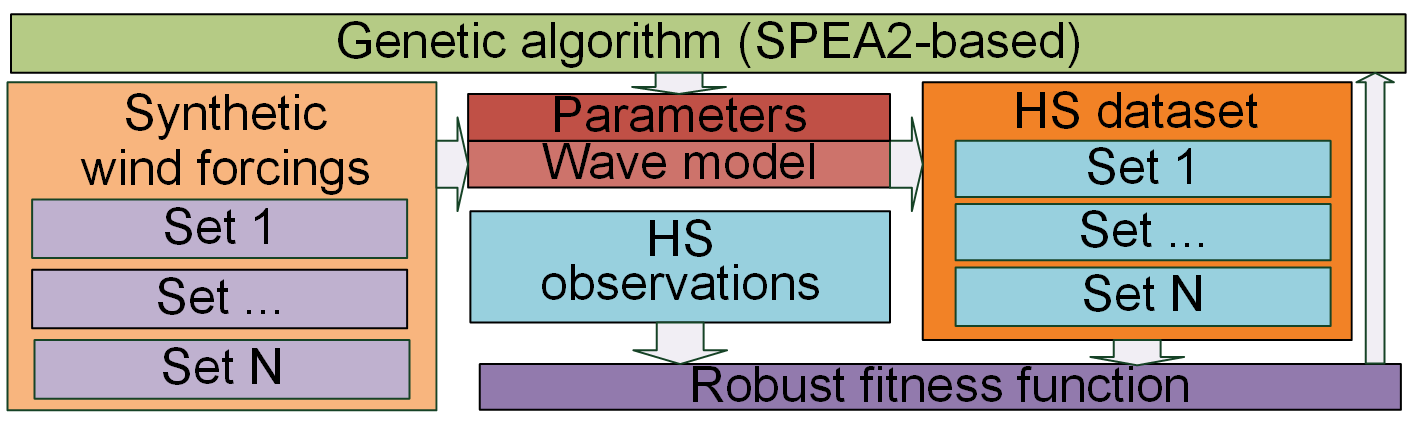}
\caption{The main logical blocks and interconnections of proposed robust evolutionary algorithm} 
\label{fig_ens_exp}
\end{figure}

For this purpose, we can adapt the baseline SPEA2 algorithm (that was introduced above) by changing the fitness assignment strategy: for a given genotype, the set of phenotypes corresponding to the elements of the ensemble is estimated and based on its values the robust metric is calculated. The flowchart of the proposed algorithm is presented in Fig~\ref{fig_ens_exp}. 

\begin{figure}
\centering
\includegraphics[scale=2.6]{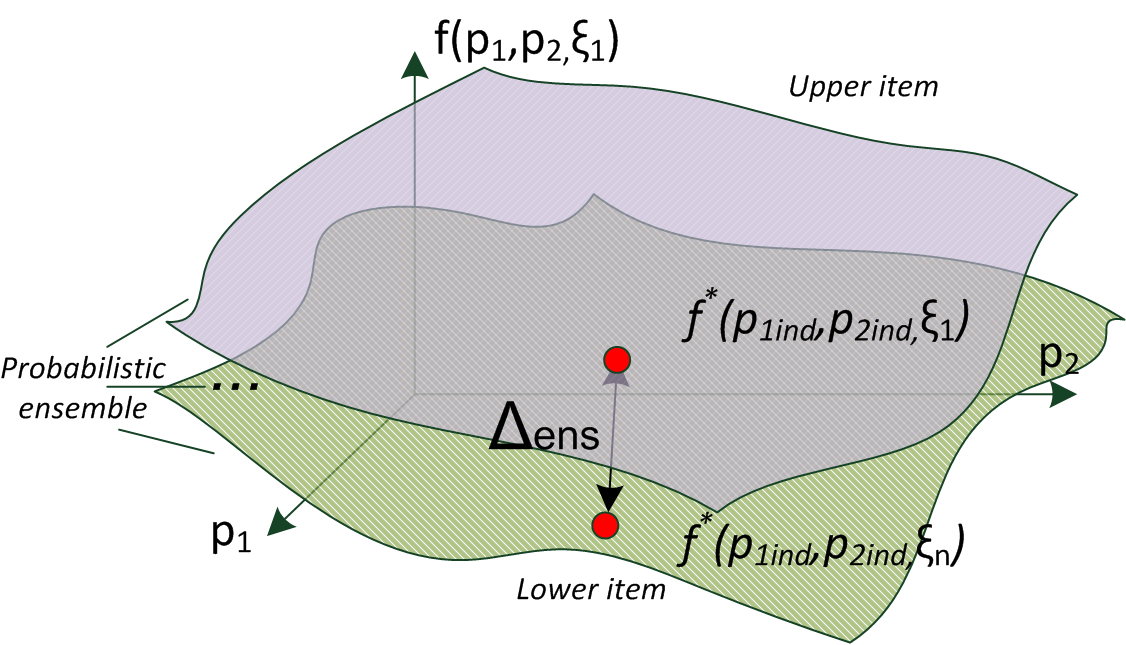}
\caption{The landscape of an objective function for a probabilistic ensemble} 
\label{fig_ens_surf_example}
\end{figure}

It is important to find a compromise between performance and robustness of the obtained solution \cite{jin2003trade}, so the fitness function for the algorithm is based on the composite estimation robustness and performance metrics. The performance can be calculated as a vector of root-mean-square errors (RMSE) against observations for a set of target points, and the robustness can be simulated in various ways \cite{mcphail2018robustness}. Fig.~\ref{fig_ens_surf_example} depicts the set of the ensemble error surfaces that are used for metric calculation. 

We tried to use the mean-variance as a robust metric, but it causes the domination of the solutions with low wind drag and, consequently, near-zero wind-induced variability. So, the ensemble mean was chosen as a trade-off metric. The pseudocode of the final implementation of the robust algorithm is presented in Alg.~\ref{alg_rspea2}.

\begin{algorithm}[H]
 \KwData{Initialised ensemble, populationSize, \\ archiveSize, crossoverRate, mutationRate}
 \KwResult{best individual from archive}
 \SetKwData{Archive}{archive}
 \SetKwData{Pop}{pop}
 \SetKwData{Union}{union}
 \SetKwData{Ind}{individ}
 \SetKwData{Model}{model}
 \SetKwData{Ens}{ensemble}
 \SetKwData{Objs}{ensObjectives}
 \SetKwData{BestObj}{bestByObjectives}
 \SetKwData{Mates}{matingPool}
 \SetKwFunction{InitPop}{InitPopulation}
 \SetKwFunction{Converged}{ConvergenceCriterion}
 \SetKwFunction{Objectives}{CalculateObj}
 \SetKwFunction{TakeBest}{TakeBestByMean}
 \SetKwFunction{Mean}{Mean}
 \SetKwFunction{Fit}{CalculateFitness}
 \SetKwFunction{Pareto}{TakeNonDominated}
 \SetKwFunction{Select}{BinaryTournamentSelection}
 \SetKwFunction{Variation}{CrossoverAndMutation}
 \Pop $\leftarrow$ \InitPop{populationSize} \\
 \Archive $\leftarrow \emptyset$ \\
 \While {not ConvergenceCriterion()}  {
    \For {\Ind in \Pop} {
        \Objs $\leftarrow \emptyset$ \\
        \For{\Model in \Ens}{
            \Objs[\Model] $\leftarrow$ \Objectives{\Ind, \Model}
        }
        \BestObj $\leftarrow$ \TakeBest{\Objs, ensAmount} \\
        \Ind.objectives $\leftarrow$ \Mean{\BestObj} \\
    }
    \Union $\leftarrow$ \Archive + \Pop \\
    \For {\Ind in \Union} {
        \Ind.fitness $\leftarrow$ \Fit{\Ind}
    }
    \Archive $\leftarrow$ \Pareto{\Union, archiveSize} \\
    \Mates $\leftarrow$ \Select{\Archive, populationSize} \\
    \Pop $\leftarrow$ \Variation{\Mates, crossoverRate, mutationRate}
 }

 \caption{The pseudocode of the implemented REBEC algorithm}
 \label{alg_rspea2}
\end{algorithm}

\subsection{Synthetic input data generation with artificial noise}

To implement the proposed probabilistic optimisation method, we developed the supplementary algorithm that allows to add specific noise to wind velocity variables ---  U (eastward) and V (northward) vector components from atmospheric reanalysis data that is used by the wave model as an external forcing. 

The algorithm starts from uniform scattering of the randomly-located sources of artificial noise in the gridded data. To obtain the realistic wind field after the application of noise, the time-spatial correlation terms are added to control the noise spreading from the source. 

The noise function for the wind vector component U produced by one noise point can be written as:

\begin{equation}
f^{*}(j,t)=N(0,\sigma ) \cdot corr(U_{j},V_{j}) \cdot corr(U_{t},U_{t-1})
\end{equation}

where j is the spatial index of source points, t is the time step index, $\sigma$ is the standard deviation parameter of the Gaussian distribution, U is the matrix of wind U-components.

Then, the aggregated noise from N source points for specific data point induced by all source points can be obtained as:

\begin{equation}
f(i,t)=\sum_{j=1}^{N}f^{*}(j,t) \cdot corr(U_{i},U_{j})
\end{equation}

where i is the spatial index of the data point, j is the spatial index of the noise point, t is the time step index, N is the number of noise points, U is the matrix of wind U-components.

\begin{figure} [ht!]
\centering
\includegraphics[scale=0.5]{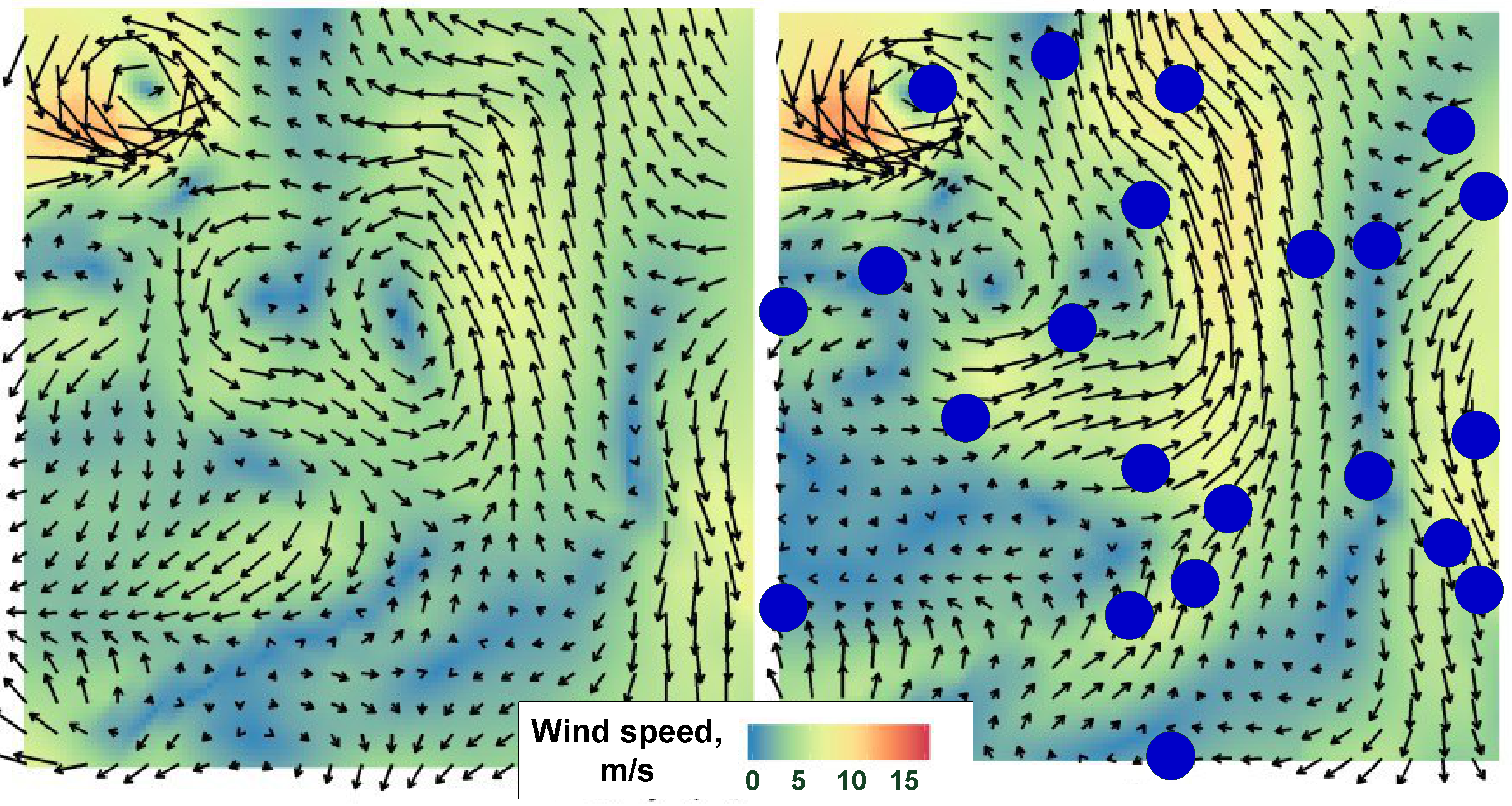}
\caption{The example of comparison of (a) basic ERA-Interim for Kara Sea region and (b) wind field augmented with noise for the same region. The blue marks depicts the noise source points locations.} 
\label{fig_wind_comp}
\end{figure}

The example of the wind field augmented with noise by the described method (with $\sigma$ equal to 25\% of basic value magnitude) is presented in Fig.~\ref{fig_wind_comp}.

It can be seen that the common wind patterns are similar but some wind speed variability exists. The additional post-processing procedure was applied to the perturbed model runs output to suppress the non-realistic wind height peaks in the observed calm periods. However, the near-peaks variability was preserved.

In this way, the ten wind data sets augmented with artificial noise were generated and used in an experimental study.

\section{Experimental study}
\label{sec_experiments}

The case study for the calibration task is based on the SWAN model configuration for the Kara Sea region. The significant wave height (Hsig) variable was chosen as a target variable. Moreover,the results in nine representative points were analyzed (P1-P9 presented in Fig.~\ref{fig_karasea_domain}) to take into account possible spatial variability of the optimal solution.

\subsection{Synthetic data for wave observations}

The wave observations data are required for the validation of the model quality and calibration algorithm effectiveness. However, such data often cannot be obtained from open data sources. To perform a reproducible experiment with Kara Sea configuration, we used the simulation results from the high-resolution WaveWatch III model \cite{tolman2009user} configuration. The systematic biases of synthetic observations against model were removed. Then we analysed the error metrics for the significant wave height variable against real observations in points №1-№3 (RMSE is 0.29m and MAE is 0.21m). We accept the quality of the WaveWatch III output as sufficient to be used as the reference dataset for the optimisation algorithms' evaluation. 

To maintain the variability of experimental scenarios, we prepared 18 subsets of synthetic observational points to be used for calibration. They consist of observation points located in various spatial areas with different depths and distances to the coast. To reproduce various scenarios, the calibration subsets were initialised with random point groups of a certain size: from a single-point situation to the all-points-instead-one case. For each subset, the points that were not used during the calibration were assigned to validation sets.

\subsection {Model configuration}

\begin{figure}
\centering
\includegraphics[scale=0.8]{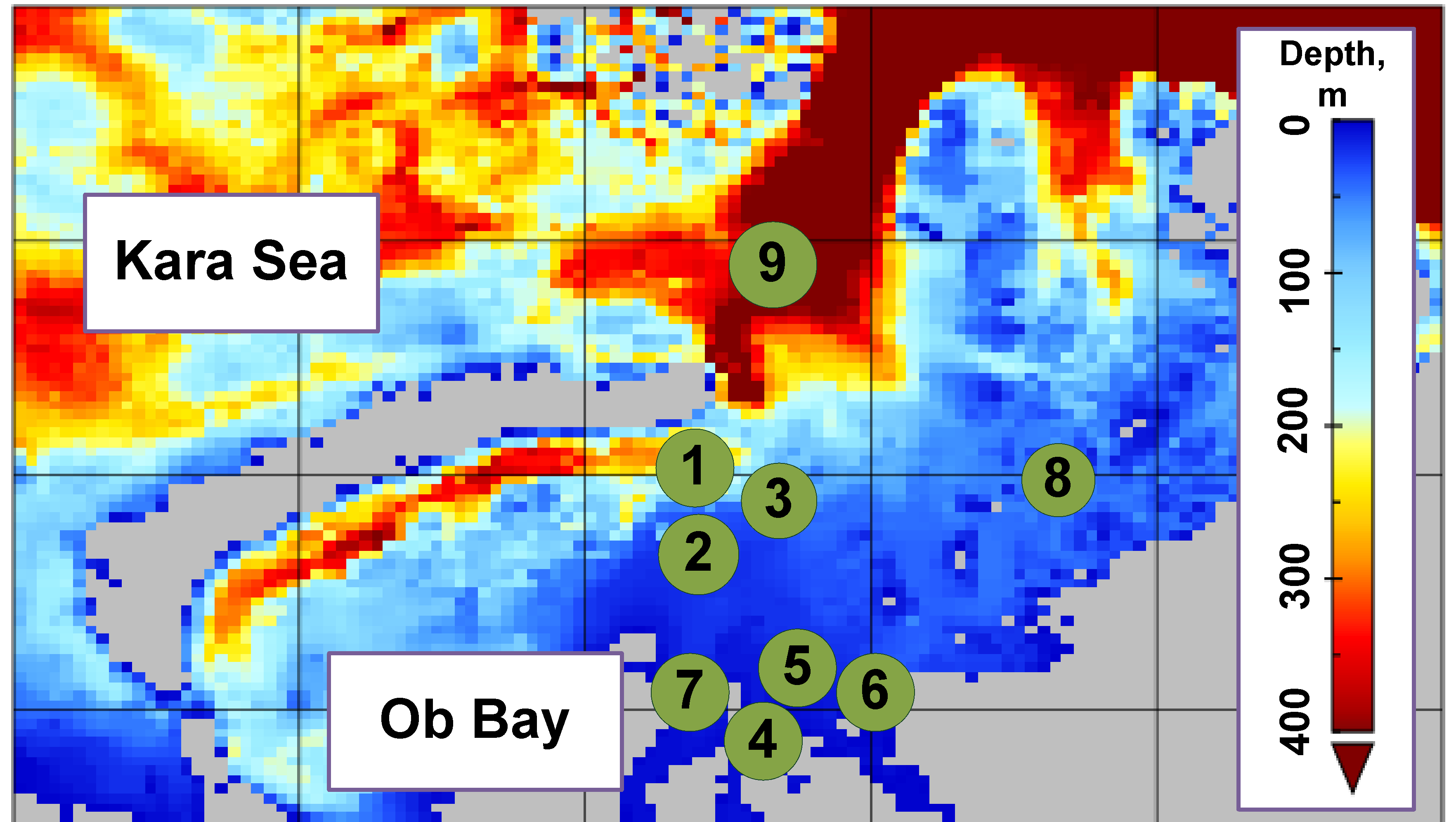}
\caption{The part of the bathymetry of the simulation domain: the Kara Sea and Ob bay. The land cells are shaded with a gray mask. The locations of observation points and their indices are specified with green marks.} \label{fig_karasea_domain}
\end{figure}

The SWAN model was configured with the regular curvilinear grid in cartesian coordinates. The initial conditions were obtained for a preliminary monthly spin-up run. The boundary conditions were not set (since the control points were distanced from the grid boundaries, see Fig.~\ref{fig_karasea_domain}). The simulation dates range was set from 20140814.120000 to 20140915.000000. The time step for integration was defined as 120 min and output time step is 3 hours. The parameterisations GEN3, COLLINS, QUADRUPL, TRIAD and DIFFRACtion were enabled. The output was configured to obtain the significant wave height (HS) values in 9 spatial points. Their locations are specified in Fig.~\ref{fig_karasea_domain}.

\subsection{Sensitivity analysis}

\begin{figure}
\centering
\includegraphics[scale=0.45]{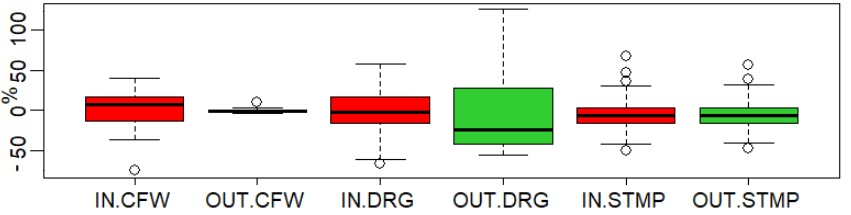}
\caption{The boxplots of relative parameter variance (IN prefix) and relative averaged RMSE of model results (OUT prefix) obtained after 50 runs. The CFW is the Collins bottom friction coefficient, DRG is the wind drag function, STMP is wave steepness. } 
\label{fig_sens_res}
\end{figure}

The sensitivity analysis of model parameters described in Sec.~\ref{sec_problem} was performed to estimate their significance. We ran the set of experiments with every parameter independently modified by additive noise with Gaussian distribution with $\sigma$/$\mu$ = 0.25 assumption. The comparison diagram of model parameters' relative variance and relative averaged RMSE metric are presented in Fig.~\ref{fig_sens_res}. The boxplots are obtained from 50 experiments with sequential noising of each variable from the chosen set.

It can be seen that the wind drag is the most sensitive parameter with a high relative output and input variance ratios, the wave steepness is the second one and the sensitivity bottom friction coefficient is quite low for most of the comparison points. In can be concluded that the SWAN model error function has a wide "plateau" with similar error values and many local minimums in the "valley" area that can affect the algorithm's convergence and robustness. 

\subsection{Validation of REBEC approach}

A set of experiments was conducted to compare the results of optimisation experiments. The initial population for both calibration approaches was produced using Latin hypercube sampling (LHS) in the parameter space. The parameters for both calibration algorithms was chosen as: population size is set to 20 individuals, the number of generations ---  60, the archive size ---  5 individuals, the probability of mutation and crossover ---  0.2.

Two objective functions were chosen for model results quality assessment: the mean absolute error (MAE) and root mean square error (RMSE).

\begin{figure}
\centering
\includegraphics[scale=0.8]{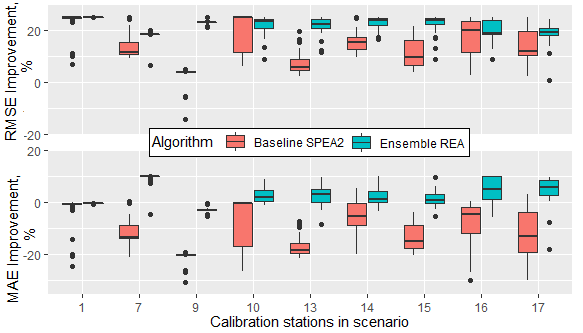}
\caption{The comparison of the baseline and robust algorithms' performance on the validation set of stations in all scenarios. The RMSE, MAE, peak-RMSE and peak-MAE metrics are presented as an improvement against the corresponding values for the default configuration.} 
\label{fig_boxplots}
\end{figure}

The calibration for every scenario was repeated 100 times to obtain the distribution of the relative improvement of RMSE and MAE against the model configuration with default parameter values (DRF=1.0, CFW=0.015, STPM=0.00302). Also, the mean relative standard deviation of the calibrated parameters set is provided. The boxplots for the scenario №3 are presented in Fig.~\ref{fig_boxplots}.

It can be seen that the variance of metrics for the robust algorithm is lower and the quality is better. The detailed metrics for all scenarios and stations sets are provided in Table~\ref{tab_results}.

\begin{table}[]
\label{tab_results}
\caption{Error metrics for the baseline and robust algorithms. The "test" block contains the metrics for the verification points. The "train" block contains the metrics for the calibration points. The boldface numbers indicate the best metrics for all station sets (the higher improvement and lower standard deviation is better)}
\begin{tabular}{cccccccccccccccc}
\hline
\multirow{4}{*}{\rotatebox{90}{Scenario}} & \multirow{4}{*}{\rotatebox{90}{Algorithm}} & \multicolumn{7}{c}{Validation points} & \multicolumn{7}{c}{Calibration points} \\ \cline{3-16} 
 &  & \multicolumn{6}{c}{Improvement, \%} & \multirow{3}{*}{\begin{tabular}[c]{@{}c@{}}Par. \\ SD\end{tabular}} & \multicolumn{6}{c}{Improvement, \%} & \multirow{3}{*}{\begin{tabular}[c]{@{}c@{}}Par. \\ SD\end{tabular}} \\
 &  & \multicolumn{3}{c}{RMSE} & \multicolumn{3}{c}{MAE} &  & \multicolumn{3}{c}{RMSE} & \multicolumn{3}{c}{MAE} &  \\ \cline{3-8} \cline{10-15}
 &  & Mean & Max & SD & Mean & Max & SD &  & Mean & Max & SD & Mean & Max & SD &  \\ 
 \hline
\multirow{2}{*}{1-9} & BL & \textbf{11} & \textbf{22.1} & 7.8 & -9.2 & 1.8 & 7.1 & 3.3 & 6.6 & \textbf{26.4} & 13.7 & 2.1 &  13.6 & 6.8 & 3.3
 \\
 & RB & 11 & 15 & \textbf{2.7} & \textbf{4} & \textbf{6.9} & \textbf{2} & \textbf{2.7} & \textbf{11.3} & 16.4 & \textbf{3} & \textbf{2.5} & \textbf{6.5} & \textbf{3} & \textbf{2.7} \\
\multirow{2}{*}{10-14} & BL & 16.6 & \textbf{23.6} & 4.9 & -7.2 & 0.7 & 5.2 & 2.7 & \textbf{24.4} & \textbf{29.1} & 4.1 & 7.9 & 12.8 & 2.7 & 2.7
 \\
 & RB & \textbf{18.1} & 21 & \textbf{2.8} & \textbf{4.1} & \textbf{9.4} &\textbf{ 2.5} & \textbf{2.4} & 22.9 & 26.9 & \textbf{3.9} & \textbf{8.7} & \textbf{14.4} & \textbf{2.2} & \textbf{2.4}
 \\
\multirow{2}{*}{15-18} & BL & 17.6 & \textbf{25.1} & 5.7 & -6.6 & 2.8 & 6.5 & 3.1 & \textbf{27.2} & \textbf{34.2} & 6.2 & \textbf{11.2} & \textbf{17.1} & \textbf{2.8} & 3.1 \\
 & RB & \textbf{18.7} & 24.7 & \textbf{4.1} & \textbf{5.4} & \textbf{10} & \textbf{4.1} & \textbf{3} & 23.4 & 33.4 & \textbf{5.1} & 11.2 & 17.9 & 3.2 & \textbf{3}
  \\
\multirow{2}{*}{All} & BL & 14.0 & \textbf{23} & 6.6 & -8.4 & 1.7 & 6.4 & 3.2 & 14.5 & \textbf{27.9} & 9.7 & 4.7 & \textbf{13.5} & 4.9 & 3.2 \\
 & RB & \textbf{14.6} & 18.3 & \textbf{3.1} & \textbf{4.6} & \textbf{8.7} & \textbf{2.7} & \textbf{2.8} & \textbf{15.8} & 21.6 & \textbf{3.9} & \textbf{5.2} & 10.5 & \textbf{2.9} & \textbf{2.8}
 \\ \cline{1-16} 
\end{tabular}
\end{table}

As can be seen, the robust approach provides a better or equivalent improvement of model performance for the validation points in all groups of scenarios. The standard deviation for both model parameters and relative improvement values are also lower than the baseline. In can be concluded that the optimal algorithm choice for validation points varies in different scenarios. The scenarios 1-9 operate with a single-point calibration set. The performance of the robust algorithm for this group of validation points is similar to baseline RMSE (but outperforms it for the MAE metric and calibration points metrics). For the other scenarios, the gain are near 1-2\% RMSE and 10\% MAE against the baseline.

Also, the calibration set quality averaged for all scenarios for the robust approach also outperforms the baseline. The standard deviation of the obtained metrics is smaller for all scenarios, as well as the mean standard deviation for model parameters. We can claim that a robust approach is effective for the cases with several spatially scattered points that can be applied for calibration. It is important to notice that the calibration points' quality is not affected in a negative way.

\section{Conclusion}
\label{sec_concl}
In the paper, the practical approach to the calibration of numerical wave models under data quality and availability constraints was proposed. The algorithm for the simulation of artificial data diversity was implemented and applied to the ERA-Interim reanalysis wind data. The regional configuration of the SWAN model was used as a case study for the parameters tuning algorithm effectiveness evaluation. 

The proposed REBEC approach was compared with the baseline SPEA2 algorithm in a set of experiments. The lower variability and better performance metrics for the spatially distributed calibration and verification points were obtained. It confirms the effectiveness of the robust calibration approach for the simulation domains with a small number and poor coverage of real observations. However, the negative impact of the proposed approach for computational performance (several simulations should be performed for each candidate parameters set) makes the robust optimisation potentially non-preferable for the model configurations with the sufficient spatial coverage of observations and high-quality atmospheric reanalyses.

The source code of the algorithms for calibration, pre- and post- processing as well as the configuration files for SWAN are available in an open repository \cite{github-sources}.

\section{Acknowledgements}
The European Center for Medium-Range Weather Forecasts (ECMWF) is acknowledged for
providing ERA-Interim surface wind data.

\bibliographystyle{splncs04}
\bibliography{references}
\end{document}